\documentclass[journal]{IEEEtran}
\usepackage{amsmath,amsfonts}
\usepackage{algorithmic}
\usepackage{algorithm}
\usepackage{array}
\usepackage[caption=false,font=normalsize,labelfont=sf,textfont=sf]{subfig}
\usepackage{textcomp}
\usepackage{stfloats}
\usepackage{url}
\usepackage{verbatim}
\usepackage{graphicx}
\usepackage{cite}
\usepackage{multirow}
\usepackage{booktabs}
\usepackage[colorlinks=true, citecolor=red, pdfborder={0 0 0}]{hyperref}
\usepackage{import}
\newcommand{\cmmnt}[1]{\ignorespaces}
\newcommand{\tbf}[1]{\textbf{#1}}

\begin{document}
\title{Real-Time Hand Gesture Recognition:\\ Integrating Skeleton-Based Data Fusion and Multi-Stream CNN
}
\author{Oluwaleke Yusuf, Maki Habib, Mohamed Moustafa
    \thanks{Oluwaleke Yusuf was with The American University in Cairo (AUC), Egypt. He is now with The Norwegian University of Science and Technology (NTNU), Norway.}
    \thanks{Maki Habib is with The American University in Cairo (AUC), Egypt.}
    \thanks{Mohamed Moustafa was with The American University in Cairo (AUC), Egypt. He is now with Amazon, Seattle, USA.}
}
\maketitle

\begin{abstract}
    Hand Gesture Recognition (HGR) enables intuitive human-computer interactions in various real-world contexts. However, existing frameworks often struggle to meet the real-time requirements essential for practical HGR applications. This study introduces a robust, skeleton-based framework for dynamic HGR that simplifies the recognition of dynamic hand gestures into a static image classification task, effectively reducing both hardware and computational demands.
    Our framework utilizes a data-level fusion technique to encode 3D skeleton data from dynamic gestures into static RGB spatiotemporal images. It incorporates a specialized end-to-end Ensemble Tuner (e2eET) Multi-Stream CNN architecture that optimizes the semantic connections between data representations while minimizing computational needs.
    Tested across five benchmark datasets---SHREC'17, DHG-14/28, FPHA, LMDHG, and CNR---the framework showed competitive performance with the state-of-the-art. Its capability to support real-time HGR applications was also demonstrated through deployment on standard consumer PC hardware, showcasing low latency and minimal resource usage in real-world settings.
    The successful deployment of this framework underscores its potential to enhance real-time applications in fields such as virtual/augmented reality, ambient intelligence, and assistive technologies, providing a scalable and efficient solution for dynamic gesture recognition.
\end{abstract}


\section{Introduction}
Hand Gesture Recognition (HGR) plays a vital role in perceptual computing by enabling computational devices to capture and understand human hand gestures using mathematical algorithms. HGR has the potential to facilitate advanced applications in domains involving human-machine interactions, human behaviour analysis, active and assisted living, virtual/augmented/mixed reality, as well as ambient intelligence. However, gesture recognition faces unique challenges due to the complex morphology of the human hand, which can adopt numerous poses and varies in physical characteristics such as size and colour among individuals. Furthermore, HGR applications often operate in challenging real-world environments characterized by occlusions, changing backgrounds, noisy inputs, and the need for real-time processing.

An HGR framework must successfully navigate these morphological and environmental challenges to satisfy the requirements of developers and end-users in real-world applications. Such requirements encompass ease of use, computational demand, hardware needs, response time, and accuracy. Hand gestures are inherently dynamic, with poses and positions changing over time, introducing a temporal dimension for accurate recognition. Thus, a sequence of hand poses must be interpreted to understand the contextual meaning of a gesture. To resolve these challenges and meet performance requirements, various frameworks have been developed for dynamic gesture recognition, each utilizing different combinations of input modalities and network architectures.

Some HGR frameworks utilize ``multi-stream networks'' by combining multiple sub-networks with distinct input channels and fusing their outputs (features extracted or class scores) for the overall network's gesture recognition output \cite{Deng2023sbm,Li2021ats,Li2023mmv,Miah2023dhg,Rehan2022lco,Song2022dhg,Li2024dab}.
On the other hand, ``multimodal frameworks'' combine multiple input modalities---including RGB, depth, skeleton, optical flow, and segmentation---to provide the network with more semantic information about the gestures. Said input modalities can be processed separately in the sub-networks of a multi-stream network \cite{Balaji2024mfh,Gammulle2021ttm,Mahmud2024qdi,Yu2021smr,Tripathi2024mfe} or combined as a unified input to a ``single-stream network'' \cite{Kopuklu2018mff}.

Whether multi-stream or single-stream, HGR frameworks have employed various (combinations of) data-driven neural network architectures, such as Graph Convolutional Network (GCN) \cite{Aiman2024abh,Li2021ats,Sahbi2021sbh,Song2022dhg}, Attention Network \cite{Balaji2024mfh,Li2021ats,Liu2021hae,Li2024dab}, and 1D/2D/3D Convolutional Neural Network (CNN) \cite{Deng2023sbm,Rehan2022lco,Lupinetti20203dh,Santos2020dgr,Kopuklu2018mff}.
Also, the efficacy of CNNs at handling spatial information is often combined with that of Recurrent Neural Networks (RNN) at handling temporal information \cite{Li2023mmv,Mahmud2024qdi,Mohammed2022a3s,Narayan2023sds,Rehan2022lco,Singh2024ddh,Tripathi2024mfe} to deal with the spatiotemporal information contained in dynamic gestures. Some frameworks have also employed exotic approaches such as Hyperbolic Manifold \cite{Chen2023hhm}, Neural Architecture Search (NAS) \cite{Yu2021smr} and Spatial-Temporal Transformer \cite{Zhao2024sts}.

However, the ultimate aim of research in the HGR domain is the development of practical, real-world HGR applications for end-users. Towards that end, most developed frameworks often prioritize maximizing performance---in terms of gesture recognition accuracy---by utilizing multimodal and multi-stream approaches which require additional, specialized hardware and increased computational complexity, respectively. These frameworks also require significant training data and data augmentation for maximal performance. Said requirements result in impractical HGR applications with higher costs, reduced user-friendliness, and longer inference times. Furthermore, aside from a few frameworks \cite{Lupinetti20203dh,Narayan2023sds,Floris2024cmp}, most developed frameworks are not being integrated into applications to demonstrate their practical utility.

The optimal HGR framework and its application would aim to minimize computational costs, eliminate the need for additional hardware, and operate in real-time, all while maintaining gesture recognition accuracy comparable to state-of-the-art (SOTA) frameworks. In recent years, the exclusive use of the skeleton modality has become prevalent to reduce computational costs. In addition, the image classification domain has developed a suite of frameworks that achieve real-time performance on resource-constrained devices. Thus, a skeleton-based HGR framework that successfully transforms the dynamic hand gesture recognition task into an ordinary image classification task would come close to being optimal.

To this end, this paper presents a skeleton-based framework for dynamic hand gesture recognition that combines data-level fusion techniques with a specialized CNN architecture. This framework effectively encodes the 3D skeleton data from dynamic hand gestures into RGB images and employs an end-to-end Ensemble Tuner (e2eET) Multi-Stream CNN architecture for the subsequent image classification. Our framework underpins a robust, lightweight, and real-time HGR application \cite{Yusuf2023doa}, distinguishing our work in addressing the challenges of dynamic gesture recognition. The contributions of this paper include the following:

\begin{enumerate}
    \item We enhance the transformation of 3D skeleton data into spatiotemporal 2D through a data-level fusion process that incorporates temporal information condensation, denoising, and sequence fitting.
    \item We develop a streamlined, fully trainable multi-stream CNN architecture that strengthens semantic interpretation by integrating multiple data representations during image classification.
    \item Our framework significantly reduces hardware and computational demands for dynamic gesture recognition, while matching or exceeding state-of-the-art performance metrics across benchmark datasets.
    \item We demonstrate the practical viability of our HGR framework by deploying a real-time application that operates efficiently on standard consumer PC hardware, ensuring accessibility and ease of use in real-world settings.
\end{enumerate}

This work builds upon previous efforts in developing a lightweight real-time HGR application \cite{Yusuf2023doa}. This paper specifically focuses on the underlying framework driving the application, situating it within related works in the HGR domain and providing detailed insights into its implementation and training.
Our framework was also extensively evaluated against SOTA frameworks using five publicly available benchmark datasets: 3D Hand Gesture Recognition Using a Depth and Skeleton Dataset (SHREC2017) \cite{DeSmedt2017st3}, Dynamic Hand Gesture 14/28 Dataset (DHG1428) \cite{DeSmedt2016sbd}, First-Person Hand Action Benchmark (FPHA) \cite{Garcia_Hernando2018fph}, Leap Motion Dynamic Hand Gesture Benchmark (LMDHG) \cite{Boulahia2017dhg}, and Consiglio Nazionale delle Ricerche Hand Gestures Dataset (CNR) \cite{Lupinetti20203dh}.

Our framework achieved performance closely aligned with, and in some instances surpassing, the state-of-the-art on the aforementioned datasets, with classification accuracies ranging from -4.10\% to +5.16\% compared to the highest benchmarks. Furthermore, the low latency of the HGR application successfully demonstrated the viability of data-level fusion for practical, real-time HGR applications.
We conducted an exploratory study with the SBU Kinect Interaction Dataset (SBUKID) to assess the adaptability of our framework to Human Action Recognition (HAR). The preliminary findings from this study indicate promising potential for extending our framework to broader applications.

The rest of this paper is structured as follows:
\autoref{sec:related-work} reviews the relevant HGR literature on skeleton modality, multi-stream networks, data-level fusion, and real-time applications;
\autoref{sec:hgr-framework} details the proposed HGR framework and its underlying components;
\autoref{sec:implementation-training} outlines the configurations in the implementation and training of the framework presented in this work:
\autoref{sec:evaluation-discussion} discusses the framework's evaluation on benchmark datasets and its performance relative to the SOTA;
\autoref{sec:hgr-application} gives an overview of the real-time application; and
\autoref{sec:conclusion-future-work} concludes the paper with some future research directions.

%% ------------------------------------------------------------------------- %%
%% ------------------------------------------------------------------------- %%

\section{Related Work}
\label{sec:related-work}
This section provides an overview of relevant research in the gesture recognition domain, crucial for understanding the principles which drove our design choices during framework development. This section covers frameworks that utilize skeleton modality for gesture recognition and those that employ multi-stream network architectures. Furthermore, we explore frameworks that apply data-level fusion of spatiotemporal information and those that go further to develop real-time applications.

\subsection{Skeleton Input Modality}
Historically, HGR frameworks primarily relied on RGB videos or depth maps. However, with the evolution of technology and the emergence of a deeper understanding of gesture complexities, recent frameworks have shifted towards using skeleton poses as an input modality. This modality, usually extracted from RGB or depth sources, provides a robust representation of hand positions and is effective at addressing issues like occlusions, complex backgrounds, and differences in individual hand morphologies. However, skeleton-based methods require offline or online preprocessing to extract skeleton data (pose estimation) from the primary sources and are vulnerable to inaccuracies during this estimation.

The utility of the skeleton modality in reducing computational complexity without compromising accuracy has led to its widespread adoption by high-performing frameworks in recent years. \cite{Liu2021hae} showcased a Hierarchical Self-Attention Network using pure self-attention mechanisms to capture spatiotemporal features while achieving competitive results with reduced computational complexity. Similarly, \cite{Peng2023aeg} presented ResGCNeXt, a lightweight GCN that combines enhanced preprocessing, BottleGroup structure, and SENet-PartAttention to achieve high accuracy with fewer network parameters.

Some frameworks have attempted to further optimize the extraction of semantic information from skeleton modality. \cite{Sahbi2021sbh} proposed a GCN method that learns graph topology and uses an orthogonal connectivity basis for node aggregation, enhancing skeleton-based HGR by mitigating overfitting with structured regularization. \cite{Aiman2024abh} introduced an angle-based GCN that improves gesture recognition accuracy by adding novel edges and designing features based on angles and distances between joints. \cite{Chen2023hhm} explored a hyperbolic manifold aware network that uses hyperbolic space to capture hierarchical features and Euclidean filters for spatiotemporal features and enhance performance without relying on dynamic graphs.

Regarding performance, \cite{Sabater2021dav} established a skeleton-based hand motion representation model utilizing pose features, a Temporal Convolutional Network, and a summarization module to achieve SOTA results in intra-domain scenarios and strong performance in cross-domain scenarios. \cite{Singh2024ddh} introduced a BiLSTM-based model with a soft attention mechanism, which significantly mitigates the challenges of intra-class and inter-class variability in gesture classes. \cite{Mohammed2022a3s} also explored a Deep Convolutional LSTM model that efficiently captures spatiotemporal features from skeleton data for real-time HGR with high accuracy and fast inference.

Our proposed framework follows this trend, capitalizing on the robust encoding capabilities of the skeleton modality towards achieving SOTA performance with reduced computational complexity. Furthermore, it is important to note that skeleton-based HGR applications inherently mitigate ethical concerns about surveillance and privacy infringement due to the minimal user-identifiable information present in skeleton data \cite{Morris2020aaa, Mucha2022bpo}.

\subsection{Multi-Stream Networks}
Many high-performing HGR frameworks adopt multi-stream network architectures integrating decoupled spatiotemporal streams, multiple temporal scales, or varied input modalities. This approach captures extensive contextual details from dynamic gesture sequences and provides the model with alternate `views' that enhance its ability to distinguish between (similar) gestures.
Multi-stream networks combine information from different sub-networks at two critical stages: feature-level fusion and decision-level fusion. Feature-level fusion combines raw features extracted by the sub-networks \cite{Balaji2024mfh,Mahmud2024qdi,Tripathi2024mfe,Deng2023sbm}, while decision-level fusion merges the classification probabilities (scores) predicted by each sub-network \cite{Mahmud2024qdi,Li2024dab,Song2022dhg,Li2021ats}. These fusion techniques enable the overall network to outperform any individual sub-network, leveraging the strengths of each to achieve superior performance.

Working with multiple modalities, \cite{Balaji2024mfh} developed a Multimodal Fusion Hierarchical Self-Attention Network that fuses features from multiple modalities (skeleton, RGB, depth, and optical flow), leading to improved accuracy with reduced computational complexity. \cite{Gammulle2021ttm} established a single-stage continuous gesture recognition framework which includes unimodal and multimodal feature mapping models for effective feature mapping between various input modalities using Temporal Multi-Modal Fusion. While \cite{Yu2021smr} proposed an NAS-based method for gesture recognition with RGB and depth modalities which uses 3D Central Difference Convolution and optimized multi-sampling-rate backbones with lateral connections between the various modalities.

For a single (skeleton) modality, \cite{Deng2023sbm} presented the Multi-Features and Multi-Stream Network for real-time gesture recognition using joint distance, angle, and position features extracted with a 1D CNN to reduce model parameters and ensure comprehensive feature extraction. \cite{Li2023mmv} introduced a Multi-View Hierarchical Aggregation Network, using a novel 2D non-uniform spatial sampling strategy for optimal viewpoints and hierarchical attention architecture to fuse multi-view features. While \cite{Miah2023dhg} proposed a Multi-Branch Attention-Based Graph and General Deep Learning Model, combining graph-based spatial-temporal and temporal-spatial features with general deep learning features from three branches.

\subsection{Data-Level Fusion}
The challenges associated with recognizing dynamic hand gestures arise from their temporal attributes. Since hand gestures are composed of a sequence of hand poses changing over time, a framework must be capable of understanding the semantic connections between these successive poses for accurate gesture recognition. This issue is exacerbated with multimodal and multi-stream networks which further require precise pixel-level correspondence across various modalities and semantic alignment between the sub-networks, respectively. Thus, the effectiveness of the framework is heavily dependent on the fusion of information from all modalities and sub-networks at the data, feature or decision (score) levels.

HGR frameworks commonly utilize online fusion at the feature and decision levels which require complex network architectures \cite{Li2024dab,Yu2021smr,Li2023mmv}, substantial training data (and augmentation), and specialized loss functions \cite{Song2018sta,Gammulle2021ttm} to achieve desired performance gains. Data-level fusion is an offline preprocessing technique where raw data from multiple sources or modalities are merged into a single representation before being processed by the network. While less commonly used, data-level fusion eliminates the aforementioned issues with other fusion methods. Temporal Information Condensation, a form of data-level fusion for dynamic gestures, involves summarizing and aggregating spatiotemporal data from gesture sequences into static images as a compact unified input representation.

For RGB modality (videos), \cite{Kopuklu2018mff} introduced Motion Fused Frames, a data-level fusion strategy that integrates temporal information into static images by appending optical flow frames as extra channels, leading to better representation of the spatiotemporal states of dynamic gestures. \cite{Santos2020dgr} also presented the ``star RGB" representation, a method for encoding temporal information by condensing dynamic gestures into single RGB images. As a consequence, after encoding temporal information, these frameworks can perform gesture classification using ImageNet-pretrained CNNs with minimal modification. \cite{Santos2020dgr} further enhanced this approach by using an ensemble of two ResNet CNNs with feature-level fusion using a soft-attention mechanism.

\cite{Lupinetti20203dh} developed a skeleton-based framework whereby dynamic gestures are converted into RGB images, such that the variation of hand joint positions during a gesture is projected onto a plane and temporal information is represented with the colour intensity of the projected points. The resulting images serve as input to an unmodified, ImageNet-pretrained CNN for gesture classification. This framework, by utilizing the skeleton modality and redefining gesture recognition as image classification, was able to minimize computational costs and serve as the foundation for real-time applications. Overall, \cite{Kopuklu2018mff,Santos2020dgr,Lupinetti20203dh} demonstrate the effectiveness of temporal information condensation in reducing the computational complexity of HGR frameworks and enhancing the real-time performance of HGR applications.

While promising, the framework developed in \cite{Lupinetti20203dh} has some drawbacks. The temporal encoding results in noisy and visually similar images, while the effectiveness of the view orientations during image generation depends heavily on the initial capture of skeleton data. Furthermore, using an unmodified CNN architecture makes it difficult for the model to extract discriminative semantic information about gestures, leading to reduced performance when tested against challenging datasets \cite{DeSmedt2017st3, DeSmedt2016sbd,Garcia_Hernando2018fph} and real-world scenarios. This framework can be improved by refining temporal information condensation, optimizing the network architecture to better handle the spatiotemporal images generated through data-level fusion, and eliminating the need for a Leap Motion sensor to capture skeleton data.

\subsection{Real-Time Applications}
As previously noted, few HGR frameworks have developed real-time applications to demonstrate their practical effectiveness in real-world scenarios. Beyond maximizing classification accuracy on benchmark datasets, gesture recognition frameworks must also minimize hardware requirements, computational costs, and inference latency. These performance considerations inevitably influence design decisions during framework development.

To this end, \cite{Lupinetti20203dh} demonstrated a real-time acquisition, visualization, and classification pipeline for gesture recognition. Their initial analysis of inference time validates their combined choice of data-level fusion, skeleton modality extracted from a Leap Motion sensor, and a regular ResNet-50 model for their framework. Conversely, \cite{Narayan2023sds,Floris2024cmp} used the lightweight MediaPipe pipeline \cite{Zhang2020mho} to extract hand skeleton data from RGB videos captured by a webcam, eliminating the need for an additional specialized sensor. Furthermore, switching from RGB to skeleton modality allowed these frameworks to leverage multi-stream networks while reducing network parameters and inference time.

The real-time application developed from our proposed HGR framework \cite{Yusuf2023doa} combines the strengths of the aforementioned frameworks and their applications. Our pipeline: \textit{(1)} extracts skeleton data from RGB webcam videos using MediaPipe, \textit{(2)} performs temporal information condensation to transform the gesture recognition into image classification, generating images from multiple view orientations, \textit{(3)} utilizes an end-to-end trainable multi-stream CNN model gesture recognition, and \textit{(4)} achieves real-time performance running on the CPU of a consumer PC.

%% ------------------------------------------------------------------------- %%
%% ------------------------------------------------------------------------- %%

\section{Dynamic Hand Gesture Recognition Framework}
\label{sec:hgr-framework}
\begin{figure*}[!t]
    \centering
    \includegraphics[width=0.75\linewidth]{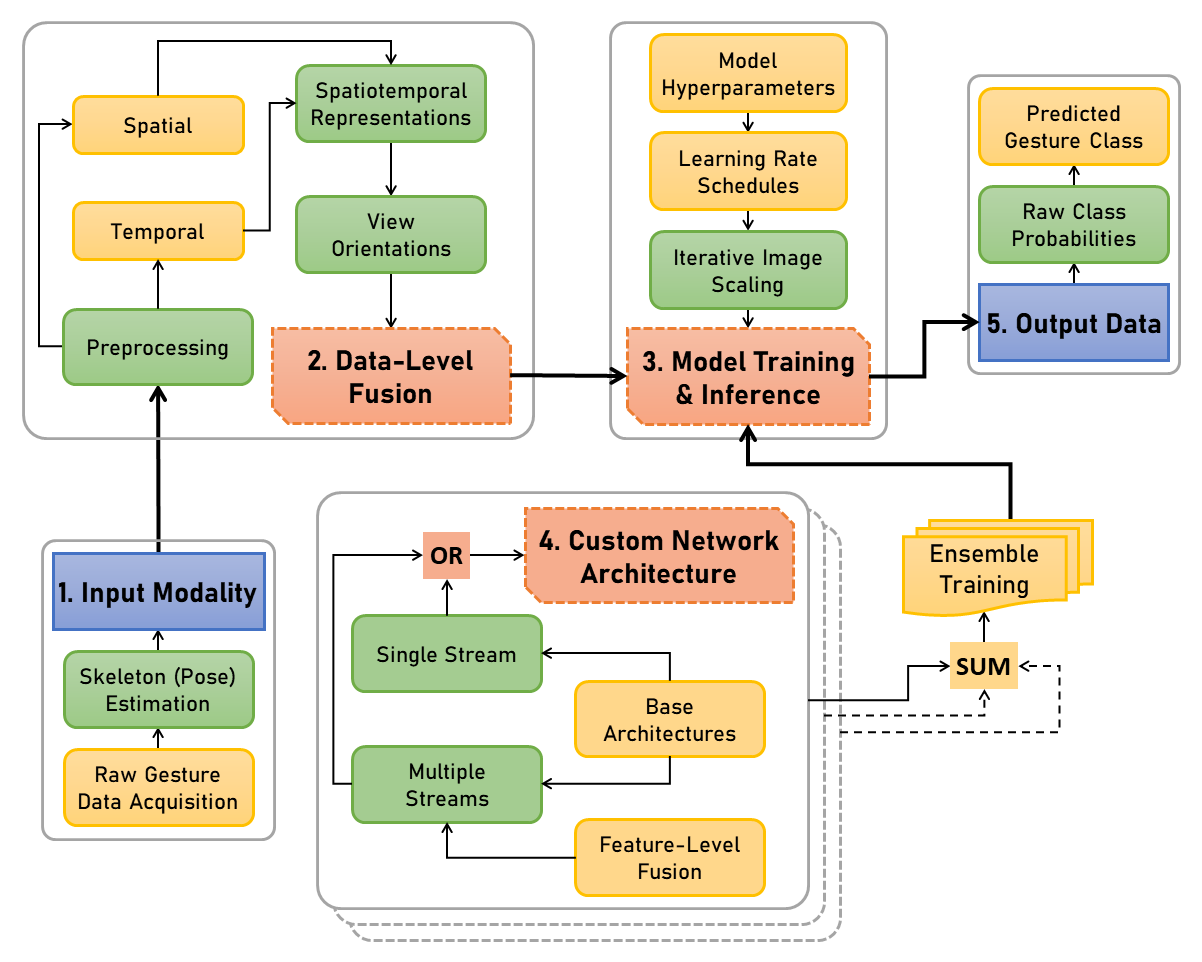}
    \caption{Diagrammatic Overview of the Proposed HGR Framework, Highlighting Key Components for Recognizing and Classifying Dynamic Hand Gestures.}
    \label{fig:proposed-framework-block-diagram}
\end{figure*}

This section elaborates on the essential elements of the proposed skeleton-based framework for dynamic HGR as shown in \autoref{fig:proposed-framework-block-diagram} as follows:
\autoref{sec:temporal-encoding} explains the method adopted for generating spatiotemporal RGB images from dynamic gesture sequences, and \autoref{sec:multi-stream-network} details the specialized network architecture designed to maximize semantic information extracted from said images. Finally, both elements are combined into a robust skeleton-based HGR framework demonstrably capable of supporting real-time applications.

\subsection{Temporal Information Condensation}
\label{sec:temporal-encoding}
Temporal information condensation, a form of data-level fusion, serves as a preprocessing stage which can be conducted offline for framework evaluation or online for real-time applications. In either case, 3d skeleton data from the decoupled spatial and temporal channels of a dynamic gesture sequence are encoded into a static 2D spatiotemporal image with a square aspect ratio. This process involves condensing the temporal information present in the original gesture sequence while retaining pertinent semantic information. As in \cite{Lupinetti20203dh}, the problem of transforming a dynamic gesture sequence into a static spatiotemporal image can be defined as follows:

If $g_i$ denotes a dynamic gesture and $S={C_{h}}_{h\text{=}1}^{N}$ represents the set of gesture sequences across $N$ classes, the temporal variation of $g_i$ is defined in \autoref{eq:gesture-variation} as follows:

\begin{equation}
    G_i = \{G_i^\tau\}_{\tau\text{=}1}^{T_i}
    \label{eq:gesture-variation}
\end{equation}

For a temporal window of size $T_i$, $\tau \in [1,T_i]$ specifies a specific timestep $\tau$, and $G_i^\tau$ indicates the frame of $g_i$ at that timestep $\tau$. Thus, the classification of dynamic hand gestures involves assigning the correct class $C_h$ to the sequence $g_i$.

The gesture sequences in the set $S$ have varying temporal windows depending on the gesture's class and its execution by individuals. Each gesture sequence $g_i$ in $S$ is resampled to a uniform temporal window $T$, set to be larger than any existing window $T_i$ in $S$. This resampling has a denoising effect which smoothens out inaccuracies in individual frames $G_i^\tau$ resulting from the pose estimation and minimizes minor variances in motion paths and sequence durations during gesture execution by individuals. Overall, denoising highlights the similarities between gesture sequences $g_i \in C_h$ in $S$, thereby improving the downstream gesture classification accuracy. Each resampled gesture $g_i$ is subsequently decoupled into its spatial and temporal channel as follows:

\begin{itemize}
    \item The \textbf{spatial} channel encodes changes in hand pose across each frame $G_i^\tau$ throughout the gesture sequence as a 3D visualization of the hand's skeleton. A distinctive palette of colours is used during visualization to create a strong contrast between the fingers and the background.
    \item The \textbf{temporal} channel encodes hand movement as a 3D visualization of ``temporal trails" of the five fingertips, from the start $G_i^1$ to end $G_i^T$ of the gesture sequence. These trails consist of a time series of markers with distinct colours corresponding to each finger. Temporal dynamics are captured by altering the transparency of these markers over time, with markers at the start of the gesture sequence ($\tau\approx 1$) more transparent and those near the end ($\tau\approx T$) more opaque.
\end{itemize}

\begin{figure*}[!t]
    \centering
    \includegraphics[width=0.89\linewidth]{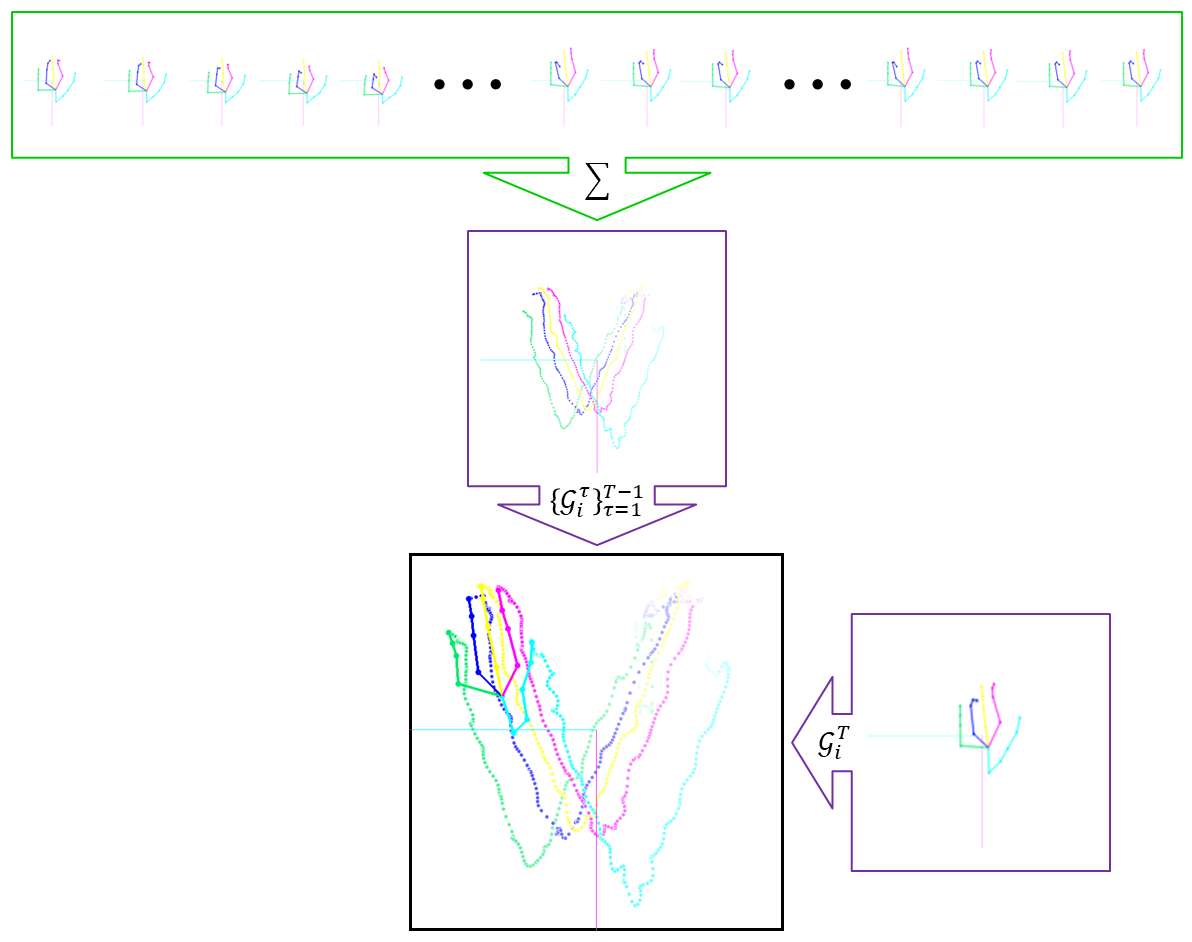}
    \caption{Temporal Information Condensation Workflow: Illustrating the Transformation of 3D Skeleton Data into 2D Spatiotemporal RGB Representations for Image Classification.}
    \label{fig:spatiotemporal-RGB-representation}
\end{figure*}

Temporal information condensation creates, for each resampled gesture $g_i$, a 3D visualization of the dynamic gesture, as illustrated in \autoref{fig:spatiotemporal-RGB-representation}. This visualization contains a compact spatiotemporal encoding of the hand pose at the final frame $G_i^T$ with the temporal trail across ${G_i^\tau}_{\tau\text{=}1}^{T\text{-}1}$ which can be projected onto any arbitrary plane (view orientation).

\begin{figure*}[!t]
    \centering
    \includegraphics[width=0.99\linewidth]{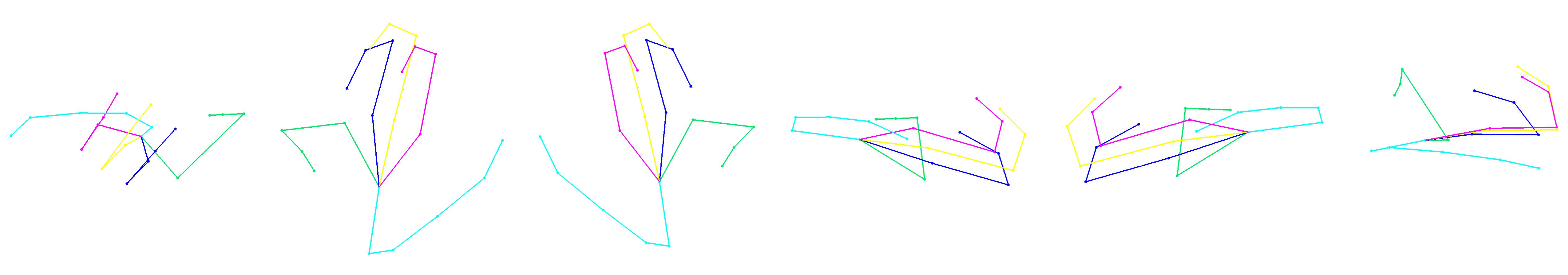}
    \caption{Illustration of View Orientations Used for Spatiotemporal Gesture Representations. Left-to-Right: Top-Down, Front-To, Front-Away, Side-Right, Side-Left, and Axonometric.}
    \label{fig:train-tap-f2s18e5-3d-all-vos}
\end{figure*}

We limit our framework to six (6) view orientations (VOs)---axonometric, front-away, side-left, top-down, front-to, and side-right---shown in \autoref{fig:train-tap-f2s18e5-3d-all-vos}. These VOs are defined by the elevation and azimuth angles of the virtual camera in the 3D visualization space, which are dataset-specific due to the varying methods used to extract the skeleton data from the raw RGB-D sources.
Thus, temporal information condensation produces a single 2D image from the 3D spatiotemporal representation of any gesture $g_i$ from any of the six view orientations $VO_j$ where $j \in [1,6]$.

However, the zoom level and position of the virtual camera need to be adjusted beforehand to ensure that the spatiotemporal representation is fully enclosed within the image, preventing any cropping or truncation. These adjustments cannot be manually configured as they differ for each gesture $g_i$ and set $S$. Instead, the sequence fitting for each gesture $g_i$ is dynamically estimated as follows:

\begin{equation}
    g_i=g_i-mean(g_i)-(L/2)
    \label{eq:center-shift}
\end{equation}

\begin{equation}
    P=(L/2)
    \label{eq:camera-position}
\end{equation}

\begin{equation}
    Z_i=max(g_i)-min(g_i)+\gamma
    \label{eq:zoom-level}
\end{equation}

\autoref{eq:center-shift} adjusts each gesture $g_i$ (a sequence of 3D skeleton coordinates) by subtracting its mean and half the target length $L$ of the static image. This adjustment shifts the gesture to the centre of the image and ensures that the virtual camera is consistently fixed at the image centre $P$ for all $g_i \in S$, as outlined in \autoref{eq:camera-position}.
\autoref{eq:zoom-level} dynamically estimates the zoom level $Z_i$ of the virtual camera for each gesture $g_i$ based on the minimum and maximum bounds of the gesture's 3D visualization. An optional padding value $\gamma$ can be added to all gestures $g_i \in S$ or set to zero.

In conclusion, temporal information condensation utilizes data-level fusion to transform dynamic hand gesture recognition into static image classification. Thus, there exists a mapping function $\Phi$ that transforms the 3D skeleton data of a dynamic gesture $g_i$ from any view orientation $VO_j$ a 2D spatiotemporal image $I_{ij}=\Phi(G_i)$. This image serves as a compact input representation for the classification task, where the goal is to determine the correct class $C_h$ for $I_{ij}$.

\subsection{Multi-Stream CNN Architecture}
\label{sec:multi-stream-network}
With the set $S$ of dynamic hand gestures $g_i$ converted into an equivalent set $S^*$ of static spatiotemporal images $I_{ij}$, traditional CNN architectures can be effectively utilized for image classification and, consequently, gesture recognition. Furthermore, transfer learning can be leveraged to enable swift convergence during model training, expedite framework development, and reduce the amount of training data and augmentation required.

For transfer learning, a base architecture is selected, and the fully connected (FC) layers of a pretrained model are customized for the new task. We evaluated ImageNet-pretrained models from various families of CNN architectures known for their SOTA performance on various benchmarks. Our selection criteria for the base architecture focused on minimizing computational footprint while maximizing classification accuracy. See \autoref{tab:cnn-architecture-analysis} for details.

\begin{figure}[!t]
    \centering
    \includegraphics[width=0.99\linewidth]{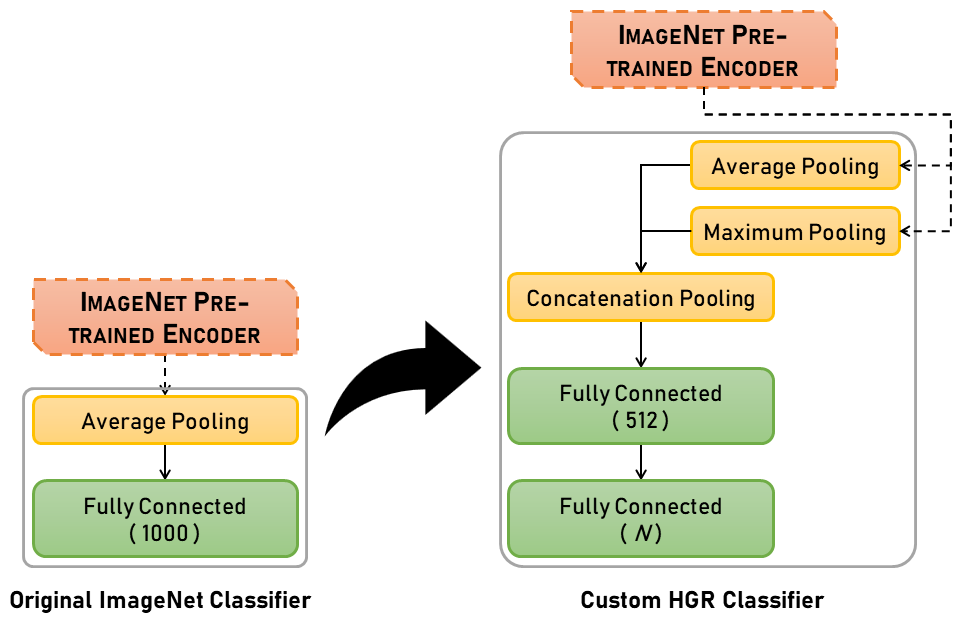}
    \caption{Comparison of Classifiers: Original ImageNet Classifier (left) versus Tailored HGR Classifier (right).}
    \label{fig:original-custom-classifiers}
\end{figure}

We replaced the original FC layers of the chosen architecture with a tailored classifier while retaining the primary convolutional layers and their pretrained weights to function as an encoder for feature extraction. Our classifier is trained from scratch and includes additional pooling, batch normalization, dropout, linear, and non-linear layers, as shown in \autoref{fig:original-custom-classifiers}.
This classifier effectively repurposes the features extracted by the encoder for gesture recognition and produces a set of probabilities, predicting the likelihood that the input image $I_{ij}$ belongs to any class $C_h,h \in [1,N]$ within the set $S^*$.

\begin{figure}[!t]
    \centering
    \includegraphics[width=0.99\linewidth]{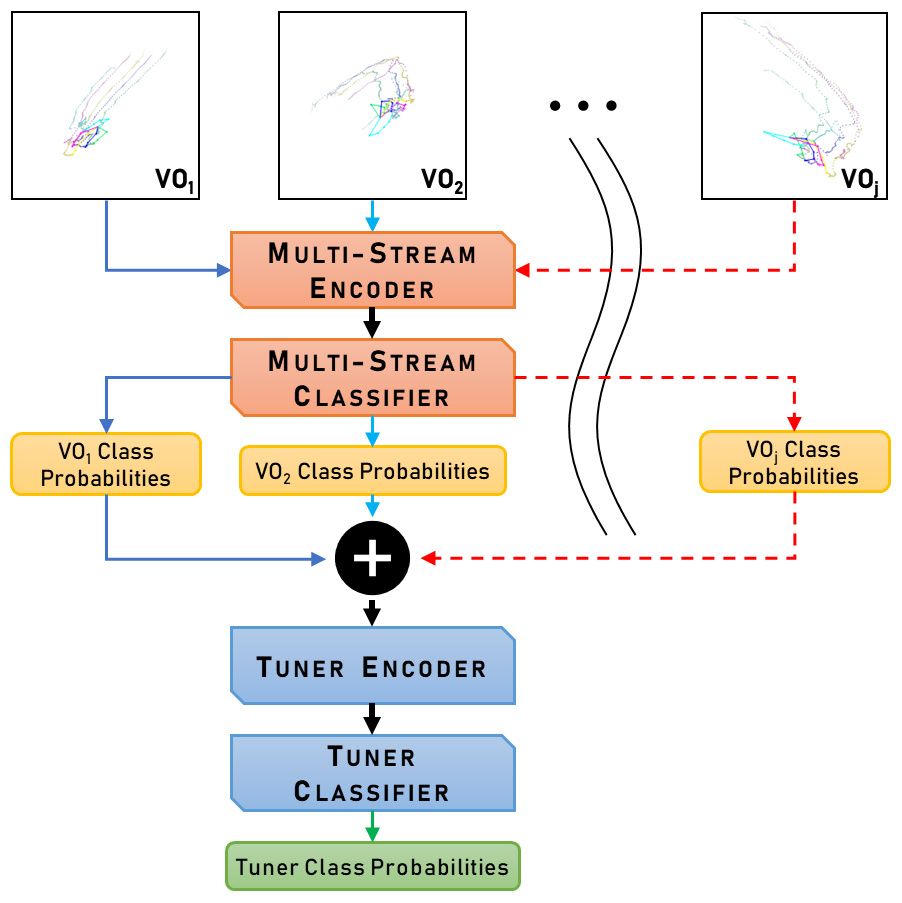}
    \caption{Illustration Showing the Processing of Static Spatiotemporal Images by the Ensemble Tuner Multi-Stream CNN Architecture.}
    \label{fig:generalized-e2eet-network-architecture}
\end{figure}

To leverage the various view orientations available for generating spatiotemporal images during temporal information condensation, we further engineered the end-to-end Ensemble Tuner Multi-Stream CNN Architecture illustrated in \autoref{fig:generalized-e2eet-network-architecture}. This architecture utilizes transfer learning and ensemble training to ensure both performance and practicality.

For any gesture $g_i \in S$, data-level fusion produces ${I_{ij}}_{j\text{=}1}^6$ spatiotemporal images from any of the $VO_j, j \in [1,6]$ view orientations available. Each image $I_{ij}$ is sequentially processed by the multi-stream encoder for feature extraction. The features are in turn processed by the multi-stream classifier in \autoref{fig:original-custom-classifiers} (right), producing a set of class probabilities ${\{CP\}_{ij}^h}_{(h\text{=}1)}^N$ for each image.

The multi-stream encoder and classifier are shared across all inputs to reduce the network's computational footprint. The multi-stream network (encoder and classifier) jointly learns the best weights for merging the view orientations guided by loss calculated for each image. Since each image yields a unique loss value, the chosen sequence (order and combination) of the input spatiotemporal images affects the performance of the multi-stream sub-network. This situation leads to an expanded search space for the optimal sequence.

The class probabilities ${\{CP\}_{ij}^h}_{(h\text{=}1)}^N$ obtained from the multi-stream classifier are transformed into a single RGB pseudo-image through online decision-level fusion. This pseudo-image is processed by the ensemble tuner sub-network which produces the final set of class probabilities. The tuner sub-network employs a lightweight version of the base multi-stream architecture's encoder while retaining a similar custom classifier.

The ensemble tuner multi-stream CNN architecture undergoes end-to-end training, yielding $(j+1)$ sets of losses and class probabilities for each gesture $g_i$. To address the homoscedastic uncertainties resulting from the encoded images and pseudo-image processed by the multi-stream and ensemble tuner sub-networks respectively, we implemented a specialized loss function inspired by \cite{Kendall2018mtl} that appropriately weighs the $(j+1)$ cross-entropy losses. However, only the final class probabilities obtained from the ensemble tuner sub-network are considered for gesture classification.

The multi-stream sub-network, by incorporating multiple inputs that are compact representations of the same dynamic gesture from different view orientations, can more accurately distinguish between gesture classes that would otherwise appear similar when visualized from a single view orientation.
In addition, using RGB pseudo-images for decision-level fusion in the ensemble tuner sub-network facilitates transfer learning and helps preserve semantic alignment among the outputs (and thus inputs) of the multi-stream sub-network.
Finally, integrating the multi-stream and ensemble tuner sub-networks into a single architecture eliminates the need to train multiple models separately, addressing a common limitation of conventional ensemble training approaches.

%% ------------------------------------------------------------------------- %%
%% ------------------------------------------------------------------------- %%

\section{Framework Implementation \& Training}
\label{sec:implementation-training}

\begin{table*}
    \begin{center}
        \caption{Virtual Camera (elevation, azimuth) Angles (in degrees) For Each View Orientation.}
        \label{tab:vo-elevation-azimuth}
        \begin{tabular}{lccccc}
            \toprule
            View Orientation & DHG1428        & SHREC2017      & FPHA          & LMDHG           \\
            \midrule
            top-down         & (0.0, 0.0)     & (0.0, 0.0)     & (90.0, 0.0)   & (0.0, 0.0)      \\
            front-to         & (90.0, 180.0)  & (90.0, 180.0)  & (0.0, 180.0)  & (-90.0, -180.0) \\
            front-away       & (-90.0, 0.0)   & (-90.0, 0.0)   & (0.0, 0.0)    & (90.0, 0.0)     \\
            side-right       & (0.0, -90.0)   & (0.0, -90.0)   & (0.0, 90.0)   & (0.0, 90.0)     \\
            side-left        & (0.0, 90.0)    & (0.0, 90.0)    & (0.0, -90.0)  & (0.0, -90.0)    \\
            custom           & (30.0, -132.5) & (30.0, -132.5) & (25.0, 115.0) & (-15.0, -135.0) \\
            \bottomrule
        \end{tabular}
    \end{center}
\end{table*}
%% ------------------------------------------------------------------------- %%
\begin{table}[tb]
    \begin{center}
        \caption{Comparative Analysis of Various Pre-trained CNN Architectures.}
        \label{tab:cnn-architecture-analysis}
        \begin{tabular}{llccc}
            \toprule
            \multicolumn{2}{l}{\tbf{CNN Architecture}} & \multicolumn{3}{c}{\tbf{Classification Accuracies (\%)}}                                                            \\
            \midrule
            Family                                     & Variants                                                 & TS1         & TS2         & Average                      \\
            \midrule
            \multirow{5}{*}{ResNet}                    & ResNet18                                                 & 80.83       & 77.62       & \multirow{5}{*}{\tbf{81.49}} \\
                                                       & ResNet34                                                 & 81.90       & 79.52                                      \\
                                                       & ResNet50                                                 & \tbf{84.17} & 80.12                                      \\
                                                       & ResNet101                                                & 82.26       & \tbf{82.86}                                \\
                                                       & ResNet152                                                & 83.10       & 82.50                                      \\
            \midrule
            \multirow{4}{*}{Inception}                 & Inception-v3                                             & 81.55       & 77.62       & \multirow{4}{*}{81.24}       \\
                                                       & Inception-v4                                             & 83.10       & 79.64                                      \\
                                                       & Inception-ResNet-v1                                      & 81.79       & 82.62                                      \\
                                                       & Inception-ResNet-v2                                      & 81.67       & 81.90                                      \\
            \midrule
            \multirow{2}{*}{EfficientNet}              & EfficientNet-B0                                          & 81.43       & 79.64       & \multirow{2}{*}{80.80}       \\
                                                       & EfficientNet-B3                                          & 81.07       & 81.07                                      \\
            % & EfficientNet-B5*                                            & 81.43 & 81.31                                 \\
            % & EfficientNet-B7*                                            & 78.93 & 81.79                                 \\
            \midrule
            \multirow{3}{*}{ResNeXt}                   & ResNeXt26                                                & 80.48       & 77.26       & \multirow{3}{*}{80.42}       \\
                                                       & ResNeXt50                                                & 82.98       & 79.52                                      \\
                                                       & ResNeXt101                                               & 82.26       & 80.00                                      \\
            \midrule
            \multirow{2}{*}{SE-ResNeXt}                & SE-ResNeXt50                                             & 81.43       & 75.95       & \multirow{2}{*}{80.33}       \\
                                                       & SE-ResNeXt101                                            & 84.05       & 79.88                                      \\
            \midrule
            \multirow{5}{*}{SE-ResNet}                 & SE-ResNet18                                              & 78.81       & 77.74       & \multirow{5}{*}{80.32}       \\
                                                       & SE-ResNet26                                              & 80.48       & 77.26                                      \\
                                                       & SE-ResNet50                                              & 81.79       & 80.24                                      \\
                                                       & SE-ResNet101                                             & 83.10       & 81.31                                      \\
                                                       & SE-ResNet152                                             & 80.60       & 81.90                                      \\
            \midrule
            \multirow{3}{*}{xResNet}                   & xResNet50                                                & 74.40       & 73.33       & \multirow{3}{*}{73.79}       \\
                                                       & xResNet50-Deep                                           & 72.26       & 73.57                                      \\
                                                       & xResNet50-Deeper                                         & 74.05       & 75.12                                      \\
            \bottomrule
        \end{tabular}
    \end{center}
\end{table}

\subsection{Benchmark Datasets}
Our proposed framework was evaluated against SOTA frameworks on the following benchmark datasets, each presenting unique challenges for the development and assessment of gesture recognition frameworks and applications.

\begin{itemize}
    \item \textbf{CNR}: The CNR dataset \cite{Lupinetti20203dh} includes spatiotemporal images captured from a top-down view, covering 1925 gesture sequences across 16 classes. This dataset is split into a training set with 1348 images and a validation set with 577 images. The absence of raw skeleton data presents a limitation as our framework cannot utilize the full capabilities of its multi-stream CNN architecture.

    \item \textbf{LMDHG}: Comprising 608 gesture sequences over 13 classes, the LMDHG dataset \cite{Boulahia2017dhg} is divided into training (414 gestures) and validation (194 gestures) sets. The dataset exhibits minimal overlap in subjects across both subsets, featuring comprehensive 3D skeleton data comprising a total of 46 hand joints for each hand.

    \item \textbf{FPHA}: This dataset \cite{Garcia_Hernando2018fph}, featuring a wide range of styles, viewpoints, and scenarios, includes 1175 gesture sequences across 45 classes. Its primary challenges stem from the similarity in motion patterns, the diverse range of objects involved, and a low ratio of gesture sequences to classes. The dataset is divided into 600 gestures for training and 575 for validation, providing 3D skeleton data for 21 hand joints per subject.

    \item \textbf{SHREC2017}: With 2800 sequences performed by 28 subjects, the SHREC2017 dataset \cite{DeSmedt2017st3} is designed for both coarse and fine-grained gesture classification, divided into 14-gesture (14G) and 28-gesture (28G) benchmarks. The dataset provides 3D skeleton data for 22 hand joints and employs a 70:30 random-split protocol for the training (1960 gestures) and validation (840 gestures) subsets.

    \item \textbf{DHG1428}: Following a structure similar to SHREC2017, the DHG1428 dataset \cite{DeSmedt2016sbd} comprises 2800 sequences performance by 20 subjects for the 14G and 28G benchmarks. It provides equivalent skeleton data and follows the same 70:30 split for training and validation.

    \item \textbf{SBUKID}: This smaller HAR dataset \cite{Yun2012tpi} contains 282 action sequences across eight classes involving two-person interaction. SBUKID provides skeleton data for 15 joints per subject and uses a five-fold cross-validation evaluation protocol, with average accuracies reported across all folds.
\end{itemize}

\subsection{Data-Level Fusion}
Our proposed HGR framework, in its generalized form, incorporates data-level fusion to create static 2D spatiotemporal images combined with the specialized end-to-end Ensemble Tuner (e2eET) CNN architecture for classifying said images. \autoref{tab:vo-elevation-azimuth} shows that the elevation and azimuth angles of the virtual camera required for the view orientations during data-level fusion are dataset-specific. The same padding value $\gamma =0.125$ was used for all datasets during sequence fitting. Note that the aforementioned settings do not apply to the CNR dataset as it only provides encoded images.

\subsection{e2eET Multi-Stream Network}
For the choice of base architecture, the performance of 26 CNN architectures from the ResNet, Inception, EfficientNet, ResNeXt, SE-ResNeXt, SE-ResNet, and xResNet families was evaluated as shown in \autoref{tab:cnn-architecture-analysis}.
ImageNet-pretrained models for said architectures were modified as described in \autoref{sec:multi-stream-network} and evaluated in single-stream mode on spatiotemporal images generated from the DHG1428 dataset with the `front-to' view orientation.
From the empirical results for two distinct training setups (TS1 and TS2), the ResNet variants ResNet-50 and ResNet-18 were chosen for the multi-stream and ensemble tuner sub-networks, respectively.

As also discussed in \autoref{sec:multi-stream-network}, the optimal sequence of spatiotemporal inputs to the multi-stream sub-network is dataset-specific due to the specifics of how the raw gestures were collected and the skeleton data extracted.
Our experiments showed that combinations of three unique VOs are adequate for reaching SOTA performance for gesture recognition. Thus, we employed the following iterative approach to determine the optimal sequence of VOs for each dataset:

\begin{enumerate}
    \item Each VO was trained separately in a single-stream network.
    \item Paired combinations of the top-performing VOs were trained in a two-stream network.
    \item Triple combinations of the top-performing VO pairs were trained in a three-stream network.
    \item The top-performing VO triplet was chosen as the optimal sequence for the dataset.
\end{enumerate}

\subsection{Model Training}
The developed framework was developed using Python 3.8.5 and tested on a Linux Ubuntu 18.04 server with four NVIDIA GeForce GTX TITAN X graphics cards.
For data-level fusion, the \href{https://vispy.org/}{Vispy} visualization library was utilized, whereas \href{https://opencv.org/}{OpenCV} was employed to generate pseudo-images for decision-level fusion.
Furthermore, the entire machine learning workflow, including the design of network architectures and the training of models, was carried out using \href{https://pytorch.org/}{PyTorch} and \href{https://www.fast.ai/}{FastAI}.
The code for this paper can be found at this GitHub repository: \href{https://github.com/Outsiders17711/e2eET-Skeleton-Based-HGR-Using-Data-Level-Fusion}{https://github.com/Outsiders17711/e2eET-Skeleton-Based-HGR-Using-Data-Level-Fusion}.

During data-level fusion, each gesture sequence was resampled to the same temporal window $T=250$. The encoded images and pseudo-images were generated with a 1:1 aspect ratio at 960px and 224px, respectively.
During the training phase, we used a batch size of 16, the Adam optimizer, and a custom homoscedastic cross-entropy loss function. The training was conducted in multiple stages, with the dimensions of the static input images progressively increasing from 224px to 276px, 328px, and 380px.
The cosine annealing schedule was applied to adjust the learning rate, with the initial rate for each stage automatically determined using FastAI's \href{https://docs.fast.ai/callback.schedule.html#learner.lr\_find}{learner.lr\_find} method.
Furthermore, various data augmentation techniques were applied during the training process, including random horizontal flips, affine transformations, perspective warping, random zooms, random rotations, and adjustments in colour space.

%% ------------------------------------------------------------------------- %%
%% ------------------------------------------------------------------------- %%

\section{Evaluation \& Discussion}
\label{sec:evaluation-discussion}
This section presents comparative evaluations of our proposed framework against other frameworks on the benchmark datasets. We also present the results of our ablation study on the efficacy of our framework in the human action recognition domain.

\subsection{Evaluation: CNR Dataset}
For the CNR dataset (\autoref{tab:cnr-comparison}), our framework had a slightly lower performance than the SOTA \cite{Lupinetti20203dh} by -1.73\%. This performance drop was due to the absence of the raw skeleton data for the CNR dataset, which prevented our framework from leveraging the enhancements in temporal information condensation and our specialized multi-stream network.

\begin{table}[h]
    \begin{center}
        \caption{Comparison of Validation Accuracy with SOTA on the CNR Dataset}
        \label{tab:cnr-comparison}
        \begin{tabular}{lc}
            \toprule
            Method                                   & Classification Accuracy (\%) \\
            \midrule
            \textit{e2eET (Ours)}                    & 97.05                        \\
            Lupinetti et al. \cite{Lupinetti20203dh} & \tbf{98.78}                  \\
            \bottomrule
        \end{tabular}
    \end{center}
\end{table}

\subsection{Evaluation: LMDHG Dataset}
For the LMDHG dataset (\autoref{tab:lmdhg-comparison}), our framework surpassed the SOTA results presented in \cite{Mohammed2022a3s} by +5.16\%, attributed to the utilization of our improved data-level fusion approach for generating spatiotemporal images. These evaluation results highlight the effectiveness of our data-level fusion enhancements and the subsequent shift from dynamic hand gesture recognition to static image classification. In addition, our transformation process effectively preserved crucial semantic information by utilizing an optimal sequence of VOs---[custom, front-away, top-down].

\begin{table}[h]
    \begin{center}
        \caption{Comparison of Validation Accuracy with SOTA on the LMDHG Dataset}
        \label{tab:lmdhg-comparison}
        \begin{tabular}{lc}
            \toprule
            Method                                   & Classification Accuracy (\%) \\
            \midrule
            Boulahia et al. \cite{Boulahia2017dhg}   & 84.78                        \\
            Lupinetti et al. \cite{Lupinetti20203dh} & 92.11                        \\
            Mohammed et al. \cite{Mohammed2022a3s}   & 93.81                        \\
            \textit{e2eET (Ours)}                    & \tbf{98.97}                  \\
            \bottomrule
        \end{tabular}
    \end{center}
\end{table}

\subsection{Evaluation: FPHA Dataset}
The FPHA dataset is particularly challenging for HGR evaluation due to the low ratio of gesture sequences (1175) to gesture classes (45). This difficulty is compounded by the 1:1 evaluation protocol, which results in closely balanced training and validation proportions. As shown in (\autoref{tab:fpha-comparison}), our framework fell short of the SOTA \cite{Sabater2021dav} by -4.10\%, even with the optimal sequence of VOs---[front-away, custom, top-down].

\begin{table}[h]
    \begin{center}
        \caption{Comparison of Validation Accuracy with SOTA on the FPHA Dataset}
        \label{tab:fpha-comparison}
        \begin{tabular}{lc}
            \toprule
            Method                               & Classification Accuracy (\%) \\
            \midrule
            Sahbi \cite{Sahbi2021sbh}            & 86.78                        \\
            Li et al. \cite{Li2023mmv}           & 87.32                        \\
            Liu et al. \cite{Liu2021hae}         & 89.04                        \\
            Peng et al. \cite{Peng2023aeg}       & 89.04                        \\
            Li et al. \cite{Li2021ats}           & 90.26                        \\
            Li et al. \cite{Li2024dab}           & 91.83                        \\
            \textit{e2eET (Ours)}                & 91.83                        \\
            Narayan et al. \cite{Narayan2023sds} & 92.48                        \\
            Rehan et al. \cite{Rehan2022lco}     & 93.91                        \\
            Sabater et al. \cite{Sabater2021dav} & \tbf{95.93}                  \\
            \bottomrule
        \end{tabular}
    \end{center}
\end{table}

\subsection{Evaluation: SHREC2017 Dataset}
For the SHREC2017 dataset, the optimal sequence of VOs---[front-away, custom, front-to]---resulted in 14G and 28G validation accuracies of 97.86\% and 95.36\%, respectively. The results presented in \autoref{tab:shrec2017-comparison} show our framework tied with the SOTA \cite{Zhao2024sts} with differences of +0.24\% and -0.47\% respectively.

\begin{table}[h]
    \begin{center}
        \caption{Comparison of Validation Accuracy with SOTA on the SHREC2017 Dataset}
        \label{tab:shrec2017-comparison}
        \begin{tabular}{lccc}
            \toprule
            \multirow{2}{*}{Method}                & \multicolumn{3}{c}{Classification Accuracy (\%)}                       \\
                                                   & 14G                                              & 28G   & Average     \\
            \midrule
            Aiman et al. \cite{Aiman2024abh}       & 94.05                                            & 89.04 & 91.72       \\
            Mahmud et al. \cite{Mahmud2024qdi}     & 93.81                                            & 90.24 & 92.03       \\
            Sabater et al. \cite{Sabater2021dav}   & 93.57                                            & 91.43 & 92.50       \\
            Balaji et al. \cite{Balaji2024mfh}     & 94.17                                            & 93.21 & 93.69       \\
            Li et al. \cite{Li2023mmv}             & 94.84                                            & 92.56 & 93.70       \\
            Liu et al. \cite{Liu2021hae}           & 95.00                                            & 92.86 & 93.93       \\
            Rehan et al. \cite{Rehan2022lco}       & 95.60                                            & 92.74 & 94.17       \\
            Peng et al. \cite{Peng2023aeg}         & 95.36                                            & 93.10 & 94.23       \\
            Mohammed et al. \cite{Mohammed2022a3s} & 95.60                                            & 93.10 & 94.35       \\
            Narayan et al. \cite{Narayan2023sds}   & 97.00                                            & 92.36 & 94.68       \\
            Deng et al. \cite{Deng2023sbm}         & 96.40                                            & 93.30 & 94.85       \\
            Miah et al. \cite{Miah2023dhg}         & 97.01                                            & 92.78 & 94.90       \\
            Li et al. \cite{Li2024dab}             & 96.90                                            & 94.17 & 95.53       \\
            \textit{e2eET (Ours)}                  & 97.86                                            & 95.36 & \tbf{96.61} \\
            Zhao et al. \cite{Zhao2024sts}         & 97.62                                            & 95.83 & \tbf{96.72} \\
            \bottomrule
        \end{tabular}
    \end{center}
\end{table}
%% ------------------------------------------------------------------------- %%
\begin{table}[h]
    \begin{center}
        \caption{Comparison of Validation Accuracy with SOTA on the DHG1428 Dataset}
        \label{tab:dhg1428-comparison}
        \begin{tabular}{lccc}
            \toprule
            \multirow{2}{*}{Method}                & \multicolumn{3}{c}{Classification Accuracy (\%)}                       \\
                                                   & 14G                                              & 28G   & Average     \\
            \midrule
            Aiman et al. \cite{Aiman2024abh}       & 90.00                                            & 88.00 & 89.00       \\
            Mahmud et al. \cite{Mahmud2024qdi}     & 90.82                                            & 89.21 & 90.01       \\
            Miah et al. \cite{Miah2023dhg}         & 92.00                                            & 88.78 & 90.39       \\
            Mohammed et al. \cite{Mohammed2022a3s} & 91.64                                            & 89.46 & 90.55       \\
            Liu et al. \cite{Liu2021hae}           & 92.71                                            & 89.15 & 90.93       \\
            Li et al. \cite{Li2023mmv}             & 92.36                                            & 89.56 & 90.96       \\
            Li et al. \cite{Li2024dab}             & 94.21                                            & 92.11 & 93.16       \\
            Narayan et al. \cite{Narayan2023sds}   & 94.64                                            & 91.79 & 93.22       \\
            Zhao et al. \cite{Zhao2024sts}         & 94.82                                            & 93.18 & 94.00       \\
            Balaji et al. \cite{Balaji2024mfh}     & 94.11                                            & 93.88 & 94.00       \\
            \textit{e2eET (Ours)}                  & 95.83                                            & 92.38 & 94.11       \\
            Li et al. \cite{Li2021ats}             & 96.31                                            & 94.05 & 95.18       \\
            Tripathi et al. \cite{Tripathi2024mfe} & 98.10                                            & 94.20 & \tbf{96.15} \\  % multimodal, remove?
            \bottomrule
        \end{tabular}
    \end{center}
\end{table}

\subsection{Evaluation: DHG1428 Dataset}
For the DHG1428 dataset, the optimal sequence of view orientations---[custom, top-down, front-away]---produced 14G and 28G validation accuracies of 95.83\% and 92.38\%, respectively. As shown in \autoref{tab:dhg1428-comparison}, our framework exhibited a marginally lower performance than the SOTA \cite{Tripathi2024mfe} by -2.27\% and -1.82\%, respectively.
While the SHREC2017 and DHG1428 datasets share similarities, DHG1428 offers an equal distribution of subjects across all classes. Thus, classification accuracies for DHG1428 14G and 28G are consistently lower in the literature compared to SHREC2017, as shown in \autoref{tab:shrec2017-comparison} and \autoref{tab:dhg1428-comparison}. Our framework follows this trend, with DHG1428 14G and 28G results being -2.03\% and -2.98\% lower than their SHREC2017 equivalents.

\begin{figure*}[!t]
    \centering
    \includegraphics[width=0.99\linewidth]{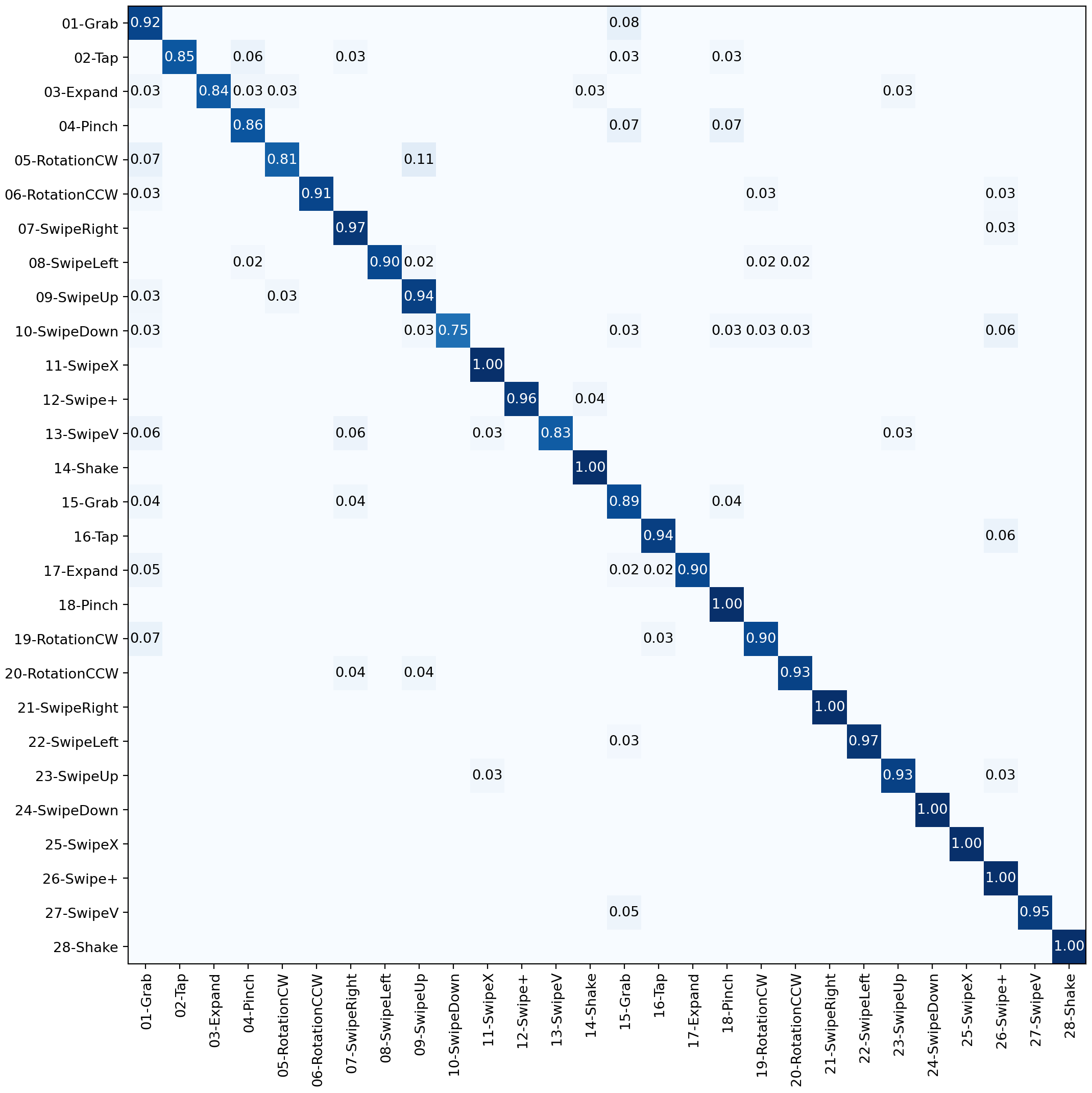}
    \caption{Confusion Matrix for the Proposed Framework on the DHG1428 28G Dataset.}
    \label{fig:confusion-matrix-dhg1428-28g}
\end{figure*}

\autoref{fig:confusion-matrix-dhg1428-28g} presents the confusion matrix of our framework’s performance on the DHG1428 28G dataset. The robust validation accuracy of 92.38\% indicates a significant alignment between the actual gesture classes and the framework's predictions. In the figure, class labels are augmented with numerical prefixes to differentiate between the DHG1428 performance modes, specifically denoting whether the gesture was executed with one finger (01-14) or the entire hand (15-28).
Consistent with the observations in previous works \cite{Chen2023hhm}, our analysis underscores a significant challenge in differentiating between the ``Grab'' and ``Pinch'' gesture classes across both 14G and 28G performance modes. This difficulty becomes evident when a visual inspection of spatiotemporal images from these two gesture classes reveals a visual similarity substantial enough to confuse the human eye.

\subsection{Ablation Study: SBUKID Dataset}
Due to the design of the data-level fusion process, we believe our proposed framework is domain-agnostic and is similarly applicable to other domains involving the classification of dynamic temporal data in the form of 2D/3D coordinates. To test this hypothesis, we adapted our framework to the skeleton-based human action recognition domain, making slight alterations to the data-level fusion process only. Similar to HGR, HAR enables computers to recognize and interpret dynamic human actions by using the entire human body as input.

We generated spatiotemporal datasets from the SBUKID dataset using all six VOs and used them to train single-stream versions of the multi-stream sub-network from our specialized e2eET CNN architecture. The evaluation results presented in \autoref{tab:sbukid-comparison} show that our framework compares favourably with frameworks designed specifically for action recognition. The best classification accuracy was obtained from the [front-away] VO, which was only -4.34\% lower than the SOTA \cite{Zhang2019van}.

This ablation study provides empirical evidence supporting the effectiveness of our data-level fusion design in transforming temporal dynamic data from various domains into spatiotemporal images for accurate classification with lower computational requirements.

\begin{table}[h]
    \begin{center}
        \caption{Comparison of Validation Accuracy with SOTA on the SBUKID Dataset}
        \label{tab:sbukid-comparison}
        \begin{tabular}{lc}
            \toprule
            \multirow{2}{*}{Method}                & Average Cross-Validation     \\
                                                   & Classification Accuracy (\%) \\
            \midrule
            Liu et al. \cite{Liu20203sg}           & 93.50                        \\
            Kacem et al. \cite{Kacem2020ang}       & 93.70                        \\
            \textit{e2eET (Ours)}                  & 93.96                        \\
            Maghoumi et al. \cite{Maghoumi2019ddg} & 95.70                        \\
            Zhang et al. \cite{Zhang2019van}       & \tbf{98.30}                  \\
            \bottomrule
        \end{tabular}
    \end{center}
\end{table}

%% ------------------------------------------------------------------------- %%
%% ------------------------------------------------------------------------- %%

\section{Real-Time Application}
\label{sec:hgr-application}

\begin{figure*}[!t]
    \centering
    \includegraphics[width=0.89\linewidth]{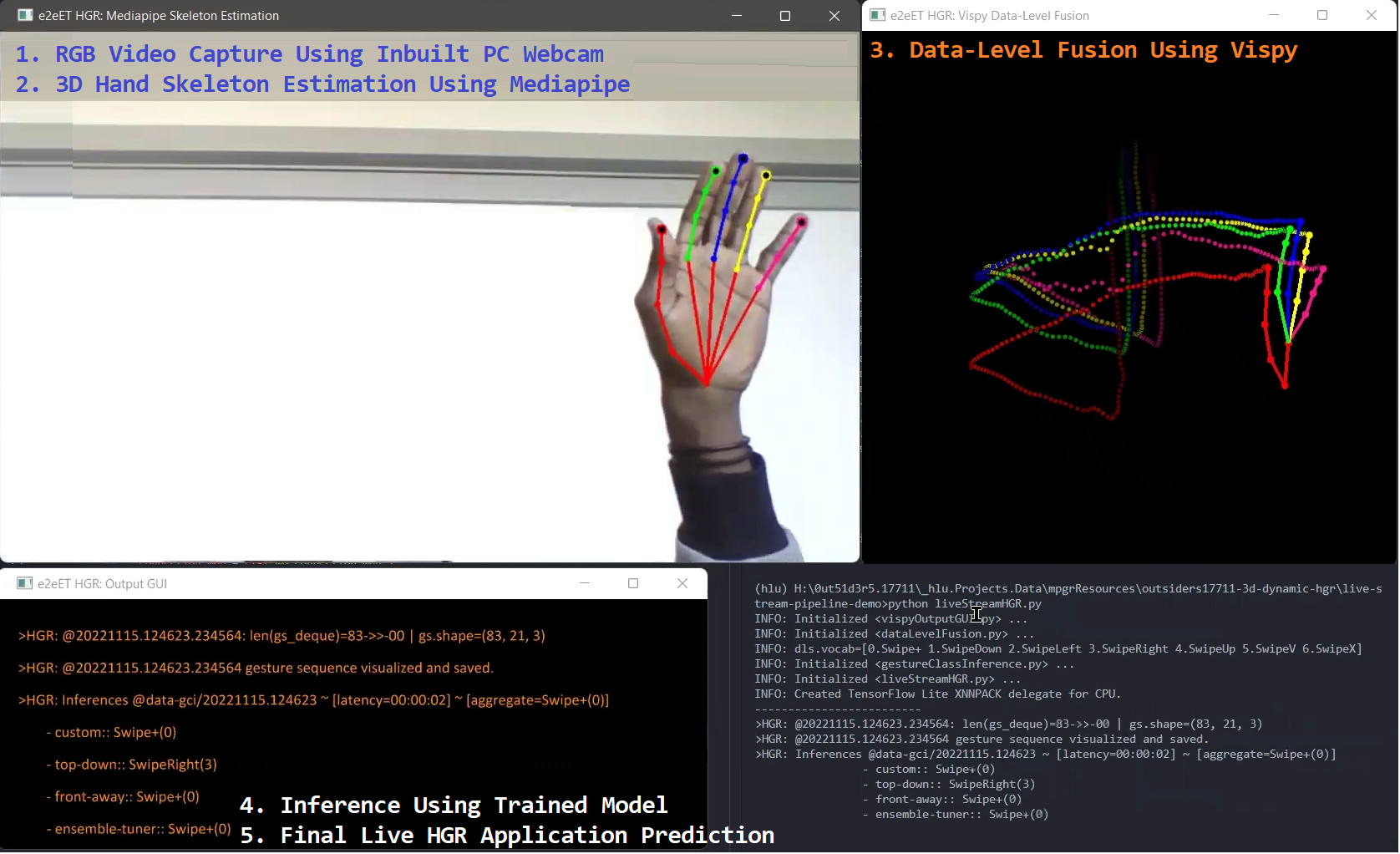}
    \caption{Demonstration of the Real-Time HGR Application, Based on Our Proposed Dynamic HGR Framework.}
    \label{fig:live-stream-demo-screenshot}
\end{figure*}

To demonstrate the practical utility of our proposed framework in reducing hardware and computational demands for HGR-based applications, we developed a real-time HGR application \cite{Yusuf2023doa} based on a generalized version of our framework. The gesture recognition in the application was powered by an e2eET multi-stream model trained on the subset of ``Swipe" gestures from the DHG1428 dataset: ``Up", ``Down", ``Right", ``Left", ``+", ``V", and ``X". The real-time operational pipeline of the prototype application is depicted in \autoref{fig:live-stream-demo-screenshot} and can be summarized as follows:

\begin{enumerate}
    \item The raw dynamic gesture data is captured as RGB videos using the inbuilt PC webcam.
    \item The skeleton data is extracted from the video frames using the MediaPipe Hands pipeline.
    \item Using data-level fusion (temporal information condensation), three spatiotemporal images are generated from [custom, top-down, front-away] VOs only.
    \item These spatiotemporal images are then processed by the trained model, which produces four class predictions: three from the multi-stream sub-network and one from the ensemble tuner sub-network.
    \item The application output---class predictions, input gesture statistics, application latency---are displayed in the graphical user interface.
\end{enumerate}

The prototype real-time application demonstrates that a standard PC webcam is sufficient for capturing raw dynamic gesture data in real-world scenarios, eliminating the need for specialized and costly sensors.
Tested on a PC with an Intel Core i7-9750H CPU and 16GB of RAM, the application captured raw gestures at 15 frames per second and maintained a latency of ~2 seconds from the user's execution of the gesture to the display of its output.
Compared to other applications running on the PC, the application's CPU and RAM consumption was acceptable, having a negligible impact on the PC’s functionality and allowing other applications to run smoothly simultaneously \cite{Yusuf2023doa}.

%% ------------------------------------------------------------------------- %%
%% ------------------------------------------------------------------------- %%

\section{Conclusions \& Future Work}
\label{sec:conclusion-future-work}

\subsection{Conclusions}
This paper explores the practical utility of existing hand gesture recognition (HGR) frameworks for real-time applications in real-world scenarios. To address the limitations posed by the considerable hardware and computational demands of these frameworks, we introduce a robust skeleton-based framework that efficiently converts the recognition of dynamic hand gestures into static image classification, while preserving crucial semantic details.

The framework incorporates an improved data-level fusion technique to generate static RGB spatiotemporal images from skeleton data of dynamic gestures and leverages a specialized end-to-end Ensemble Tuner (e2eET) Multi-Stream CNN architecture for classification. This architecture incorporates semantic information from multiple view orientations for accurate image classification while maintaining computational efficiency.

The effectiveness and generalization of the proposed framework were extensively evaluated on five benchmark datasets: SHREC'17, DHG-14/28, FPHA, LMDHG, and CNR. The results demonstrated its competitive performance, with accuracies ranging from -4.10\% to +5.16\% compared to current state-of-the-art benchmarks. Furthermore, exploratory ablation studies in the human action recognition domain demonstrated the framework's robust capability in processing temporal dynamic data across varied applications.

The practical utility of the framework was underscored by the development of a real-time HGR application, which operates with standard PC hardware using a built-in webcam and maintains low resource (CPU \& RAM) consumption. This successful implementation showcases the potential of data-level fusion to substantially reduce hardware and computational demands without sacrificing performance, making it a viable solution for real-time dynamic gesture recognition across multiple domains.

\subsection{Future Work}
Expanding our skeleton-based approach to human action recognition presents a logical progression, leveraging similarities between HGR and HAR to categorize human actions using skeletal data.
Enhancing the multi-stream network architecture by integrating attention mechanisms will eliminate the need for dataset-specific optimal sequences of view orientations and could also improve overall performance.
Further efforts should also concentrate on optimizing computational efficiency through tailored ML/DL optimization methods, enhancing both the performance and effectiveness of our framework.

Testing the framework and its applications in real-world scenarios, especially in healthcare or virtual reality, is crucial for gathering insights into its practicality and identifying areas for refinement.
A comprehensive evaluation of user experience is equally important, as it will provide essential feedback on usability that will shape future iterations of the framework. These steps will not only advance dynamic hand gesture recognition but also deepen our understanding of its applications and improve user-centric development.

%% ------------------------------------------------------------------------- %%
%% ------------------------------------------------------------------------- %%

\bibliographystyle{IEEEtran}
\bibliography{bibliographyMJM}

\end{document}